# MLP Based Continuous Gait Recognition of a Powered Ankle Prosthesis with Serial Elastic Actuator

Yanze Li, Feixing Chen, Jingqi Cao, Xingbang Yang*

*Abstract*—Powered ankle prostheses effectively assist people with lower limb amputation to perform daily activities. High performance prostheses with adjustable compliance and capability to predict and implement amputees' intent are crucial for them to be comparable to or better than a real limb. However, current designs fail to provide simple yet effective compliance of the joint with full potential of modification, and lack accurate gait prediction method in real time. This paper proposes an innovative design of powered ankle prosthesis with serial elastic actuator (SEA), and puts forward a MLP based gait recognition method that can accurately and continuously predict more gait parameters for motion sensing and control. The prosthesis mimics biological joint with similar weight, torque, and power which can assist walking of up to 4 m/s. A new design of planar torsional spring is proposed for the SEA, which has better stiffness, endurance, and potential of modification than current designs. The gait recognition system simultaneously generates locomotive speed, gait phase, ankle angle and angular velocity only utilizing signals of single IMU, holding advantage in continuity, adaptability for speed range, accuracy, and capability of multi-functions.

## I. Introduction

Nowadays, there are over 600,000 lower limb amputees in the United States, each year with estimated 100,000 people undergoing amputation of a lower limb, especially below the knee [1]. These subjects often suffer greatly from physical and mental pain due to low mobility, severely preventing them from recovery and rejoining the society. Research in robotic prosthetics has the goal to assist amputees by developing powered prostheses which mimic a healthy ankle-foot and replicate the non-pathological gait. Previous analysis has shown that a functional ankle outputs substantially 40% of power during daily activities, more than any other lower limb joint [2], and ankle moment contributes linearly to propulsion [3]. This puts forward challenging demand on replacement of power generation at the ankle by prostheses [4], which requires exact mimicking of kinetic features and torque outputs. With the purpose of reducing the metabolic cost of amputees, lower limb prosthetics evolved from rudimentary passive feet [5] to modern bionic feet with certain features on stability and propulsion [6]. The propulsive devices usually contain pneumatic [7] or hydraulic actuators [8], or motorized drivers connected to a transmission system [9], whose actuators can be either rigid or elastic depending on their mechanical structure.

actuators (SEA [11] [12], including parallel elastic actuators (PEA) [13] and other variable stiffness elastic actuators [14]). Theoretically, these technologies add an intrinsic impedance to the mathematical model of the system via either algorithm or hardware structures, introducing stiffness, damping, and inertia features [15]. Most algorithms are infeasible to reduce inherent high inertia and high friction of actuators [10] and requires extra complexity to the system and complicated adjustment via reinforcement learning or transfer learning [16],[17]. QDD relies on a proprioceptive sensor to measure the position to anticipate the torques for precise control, which increases the complexity of the system and the electrical design. Comparatively, SEA possesses higher torque density and lower heat thanks to much lower current regime than QDD [18], and also overcomes the backdrivability limitation [19],[20]. Current SEAs utilize planar torsional springs to realize a linearly alterable stiffness through different layers of stacking [21]. Drawbacks of current spring structures include excessive mass and weight, less stability under radial or axial forces, inappropriate stiffness and rigidity for prostheses, short endurance, and low potential of modifications [22] [23].

Gait recognition and analysis is also crucial for lower-limb prostheses. A common method to estimate the gait phase is to use a finite state machine and segment the gait phase into several discrete events, where a specific assistive effect is actuated, and transitions through the state machine is triggered by various sensors [24] [25]. Since each state is represented discretely instead of continuously, transitions between each state are not seamless, and distinctive assistance controllers must be designed for each state. Another widely used method is incorporating a phase variable approach with iconic parameters measured by sensors[26]. For example, a single IMU mounted on the shank can compute the gait phase and velocity[27]. This phase variable approach is more adaptable than other methods since it maps input configurations into gait phase directly, but it still requires a rhythmic motion of the subject and is feeble to accommodate abrupt changes in walking speeds [28]. For better understanding of gait dynamics, machine learning (ML) has been proved feasible with numerous works optimizing the robustness of their system via ML. Neural networks have widely been used in deep learning tasks, where they perform outstanding capability to learn difficult and complicated tasks [27] [29]. In addition, well-trained neural networks can calculate at real time once embedded into the control board and implemented as a series of matrix operations, which makes it suitable for assistance across larger speed range and continuous gait recognition that is scarcely seen in prosthetic robotics before.

This paper proposes a new design of powered ankle prosthesis inspired by the OSL project[21][30], with

Yanze Li is with the School of Automation and Electrical Engineering, Beihang University, Beijing, China (e-mail: liyanze531@126.com).
Feixing Chen, Jingqi Cao, and Xingbang Yang are with the School of Biological Science and Medical Engineering, Beihang University, Beijing, China.
Corresponding authors. (e-mail: yangxingbang@buaa.edu.cn).

innovative design of planar torsional spring utilized in SEA, and continuous gait recognition and control method using neural network with the configuration of multi-layered perceptron (MLP). Finite element analyses (FEA) are conducted to discover the relationship between stiffness and endurance with mechanical characteristics, and verify the spring's merits in stiffness, endurance, and potential of modification. In addition, individual gait data are gathered via motion capture system, and the hypothesis that utilizing single sensor to estimate locomotive speed and gait phase, and hence generate real-time output of kinetic configurations is testified via simulation and on the proposed prosthesis. The result shows our advantage in accuracy and a wildly adaptive capability to speed range (0-4m/s). This study establishes the connection between single sensor input and joint outputs via machine learning, potentially to be adopted to other wearable robotics.

## II. MECHATRONIC DESIGN

### A. Mechanical Design

The proposed prosthesis (Fig. 1) utilizes the combination of motorized driver with a belt-driven transmission system in S3M standard, which consists of two pulleys connected respectively to the motor and the SEA, and a tensioning wheel mounted on two eccentric bases that attaches to the outer housing, providing multiple choice of tension. Two pieces of outer housings collectively fix the inner transmission system while hold the motor in position with a supportive arch. Standard connectors for prosthetics are provided on both extremes of the prototype, allowing attachments to artificial feet and sockets. Two half-open chambers are placed respectively in the front and rear for storage of battery and control board, with an inner trench that connects two chambers for wiring. Detachable plastic lids that fit to the housings are made to cover the chamber, which ensures protection during operation as well as convenience for maintenance. Detailed specifications are available in Table I.

### B. Electronical Design and Control System

The prototype is powered by a LiPo battery, and uses a DC motor (Cubemars AK10-9, China) integrated with a planetary gear transmission with 9:1 speed-reduction ratio. The onboard sensors include one inertia measurement unit (IMU, Yahboom CMP10A, China) and a 3-axis loadcell sensor (HUILIZHI LZ-SWF58, China) that measures forces at the top end of the prosthesis. The integrated encoder inside the motor can retrieve parameters including but not limited to position, rotational speed, temperature, and current.

A traditional hierarchical control architecture containing three levels [31] is utilized in the controller design. A STM32 board (Robomaster A, China) is responsible for the high and middle-level control, which collects data from sensors and recognizes amputees' intended movements. A driver chip integrated in the motor manages the low-level control, generating appropriate control law based on the recognized intent and producing desired output command. Three AD conversion chips are connected to the loadcell and main control board, realizing the retrieval of force data. All communications are through USART protocol.

TABLE 1 DESIGN SPECIFICATION OF THE PROSTHESIS

| Metric | Prosthesis | Biological ankle |
|---|---|---|
| Inner reduction ratio | 4.6:1(83:18) | |
| Motion Range | ±50° | -50° (plantar flexion)~ 20° (dorsiflexion) [32] |
| Mass (kg) | 2.9 | 0.048*Bodyweight [33] |
| Bus Voltage (V) | 24 | |
| Battery Capacity (m Ah) | 3000 | |
| Continuous Torque (Nm) | 82.8 | |
| Peak Torque (Nm) | 220.8 | 200 [34] |
| Peak Rotational Speed (rad/s) | 5.65 | 3.5 [35] |
| Motor torque Constant (Nm/A) | 0.16 | |

### C. Electronical Design and Control System

The planar torsional spring (Fig. 1d) used in SEA needs to be both elastic and endurable [36], which requires a linear elasticity for easier control and lower von Mises Stress (VMS) for longer endurance. In addition, higher modifiability is preferred, meaning that a range of stiffness can be achieved by adjusting certain parameters. Based on the joint torque simulation [11], a total stiffness of approximately 20Nm/° is desired, which is realized via axially stacking of springs (Fig. 1c). In virtue of larger yield strength and longer fatigue life, aged maraging steel 300 is selected, which excels in performance of fatigue behavior [37].

## III. GAIT RECOGNITION AND CONTROL METHOD

### A. Gait Modelling and Motion Capture

Musculoskeletal models of lower limbs (Fig. 2b) at various speeds are established in OpenSim software (Version 4.4, Stanford) [38][40], from which kinetic information of the tibia can be extracted. A three-dimensional, 23 degree-of-freedom model of the human musculoskeletal system is used for gait analysis, namely the Gait 2392 Model [41]. The data of walking, including speed of 2m/s, 3m/s, 4m/s, and 5m/s are from previous work of the OpenSim team [42]. The models can broadly reflect the average gait of humans, and are imported into the neural network for rudimentary training. Gait data of the subject for prosthesis field tests were acquired by motion capture system and added to better suit individual gait habits. The Vicon MX (Vicon Motion Systems Ltd, Oxford, UK) is utilized in the research, with 10 cameras tracking the position of 18 reflective markers attached to the subject's anatomical reference points at 100Hz (Fig. 2a), and images are processed in Nexus software (version 2.6). The lower-body plug-in gait model was used to record the kinetics of the subject, which is then exported and processed by OpenSim [43]. Gait data at speeds of 0.5m/s, 0.8m/s, 1.25m/s 1.5m/s, 1.8m/s, 2m/s, 2.5m/s, 2.8m/s, 3m/s, 3.3m/s, 3.5m/s, 4.5m/s are gathered on a running machine (Fig. 2a).

### B. Neural Network

The inputs of the neural network are angles ($\theta, \varphi, \psi$) and angular velocities ($\dot{\theta}, \dot{\varphi}, \dot{\psi}$) respectively in the sagittal plane, coronal plane, and transverse plane (Fig. 3a). Since inputs of the neural network are six uncorrelated variables from IMU, configurations like convolutional neural network (CNN) that emphasizes spatial relevance in neighboring dimension is unsuitable. The intermediate variables are gait phase ($p$) and locomotive velocity ($v$), while the outputs are ankle angle ($\alpha$) and angular velocity ($\dot{\alpha}$). The whole data set took a 5-fold cross validation to gain a stable result, where one group is chosen as the test set while the others as train set in each round. Additionally, 30% of the train set are randomly separated to form the validation set.

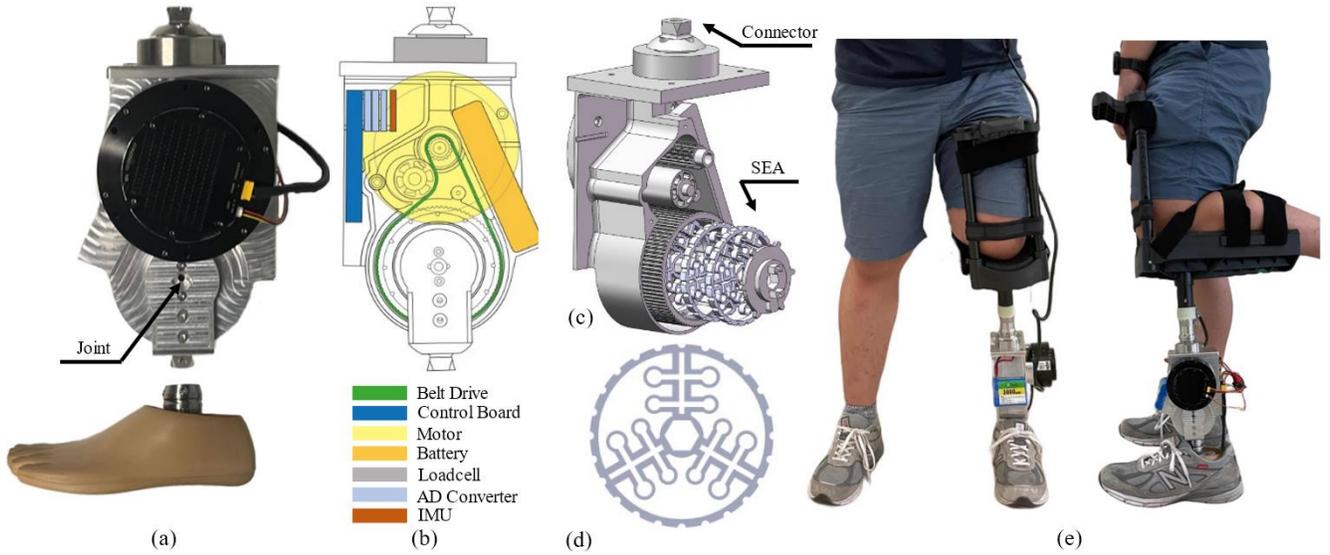

Fig. 1 Prosthesis design. (a) Physical implementation. (b) Schematics. (c) Exploded view of stacking of planar torsional spring of the SEA. (d) Planar torsional spring. (e) Subject wearing the proposed prosthesis.

The proposed MLP configuration has 27 layers, consisting of an input layer, 12 fully-connected hidden layers, each following by a rectified linear unit (ReLU) layer, and 2 output layers, one as the middle outputs, the other as the final outputs. Detailed specifications are shown in Table 2. For each hidden layer and its corresponding activation layer, the calculation function could be presented as follows:

$$\begin{cases} a_k^{n \times 1} = W_k^{n \times m} x_{k-1}^{m \times 1} + b_k^{n \times 1} \\ y_k^{n \times 1} = \sigma_{relu}(a_k^{n \times 1}) \end{cases}$$

where $x_{k-1}^{m \times 1}$ represents the output of $k-1$ layer, $W_k^{n \times m}$ means the weight metric of $k$ layer, $b_k^{n \times 1}$ stands for the bias vector of $k$ layer, $a_k^{n \times 1}$ shows the linear output of the fully-connected layer, $\sigma$ indicates the activation function, and $y_k^{n \times 1}$ is the output of $k$ layer. The evaluation methodology incorporates key metrics, namely mean square error (MSE), root mean square error (RMSE), and mean absolute error (MAE). These metrics are essential for assessing performance as they quantify the disparities between predicted values $\tilde{y}_i$ and actual values $y_i$. To ensure robust evaluation, we employed a 5-fold cross-validation approach, resulting in consistent performance fluctuations within a similar range. Consequently, the averages for each of the evaluation metrics were calculated.

After training for 200 epochs, we observed a substantial reduction in loss for both middle and final outputs. The loss for the middle output reduced from 3286.444 to 2.588, and for the final output, it reduced from 38593.39 to 2784.606. Similarly, the test data got a loss of 15.294 for the middle output loss, and 3072.222 for the final, showing no overfitting.

To further evaluate the results, we also calculated RMSE and MAE of these test data. These metrics provide insights into the model's performance in a more interpretable way than MSE, as they are in the same units as the target variables. Each dimension is evaluated separately to better interpret the learning effects. And the average RMSE and MAE values for each dimension are as follows:

For v: RMSE = 0.443, MAE = 0.331.
For p: RMSE = 2.610, MAE = 1.2387.
For α: RMSE = 5.059, MAE = 3.886.
For ά: RMSE = 53.089, MAE = 37.041.

The metrics indicate superior results in gait recognition. Further experiments and analysis are conducted in section IV B.

*C. Motor Control*

With predicted locomotive speed and gait phase, the neural network then determines the desired output angle that is actuated by the motor. A PID controller is utilized in motor control (Fig. 3b), with the encoder retrieving and feedbacking the real time position (angle) of the motor. The method guarantees an adaptable torque output with a steady kinetic output depending on the actual application.

TABLE 2 SPECIFICATIONS OF THE NEURAL NETWORK

| | |
|---|---|
| Structure | IL – 6 HL – MOL – 6 H – FOL[*] |
| Initial Learning Rate | 1e-2 |
| Criterion | MSEloss |
| Optimizer | Adam |
| Weight Decay | 1e-2 |
| Epochs | 200 |
| Batch Size | 512 |
| Activation Function | ReLU |

[*]IL: Input Layer, HL: Hidden Layers, MOL: Middle Output Layer, FOL: Final Output Layers.

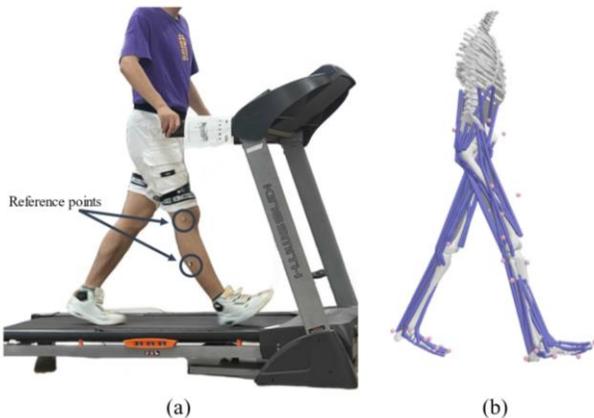

Fig. 2 Motion capture. (a) Subject experimenting on a running machine. (b) Musculoskeletal model in OpenSim

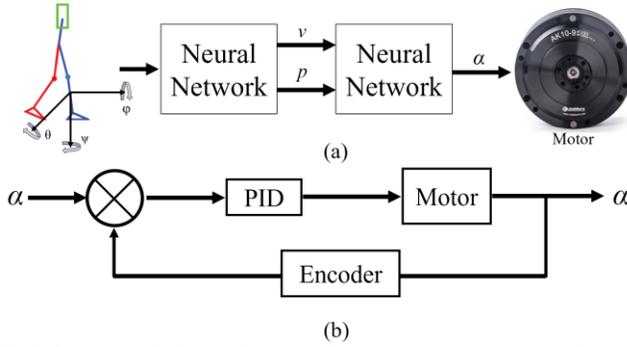

Fig.3 Control method. (a) gait recognition system. (b) Motor control

## IV. EXPERIMENTS AND RESULTS

### A. Planar Torsional Spring

Simulation and analysis of stiffness and maximum VMS were conducted in ANSYS Workbench 23.0, where the mesh size was set at 0.3 mm for convergence, leaving a margin of error of 3%. Cylindrical support was neglected due to its minute effect (below 1%) on the simulation. A fixed support was loaded onto the inner ring, while a moment of up to 20 Nm was loaded onto the outer ring to simulate the actual applicational scenarios of SEA (Fig. 4c) [36].

According to preliminary experiments, the radius of the outer circle ($R_{outer}$), thickness of the inner circle ($T_{inner}$), the centerline of the pair arm (L), and the width of the branch (d) were identified to be crucially influential among all structure parameters of the SEA (Fig. 4b). It was also proved the stiffness remains constant regardless of the variation of torque under elastic limit.

Single Factor Experiment was conducted to discover the statistical impact of each factor. Control value for $R_{outer}$, $T_{inner}$, L, and d were set at 4mm, 1mm, 9mm and 1.5mm respectively, while single variants were altered. Shapiro-Wilk tests were employed for all the data on account of small sample size [44] [45], which suggest that stiffness versus four variants were all normally distributed, yet maximum VMS did not follow normal distribution. Therefore, the Pearson's test [46] was conducted, which proved the linear relationship between the stiffness and each factor. The Spearman's test [47] proved the negative correlation of $R_{outer}$, T, and d on maximum VMS. An orthogonal experiment was then performed to optimize the design for a moderate elasticity and low VMS. The result is shown in Table 3. The stiffness of the final structure is 7.5Nm/°, and the maximum VMS is 972MPa (Fig. 4d), about half of the yield strength.

With sufficient analysis. The planar torsional spring achieves the desired mechanical properties, yet can be replaced or altered in easy operation. The selection of SEA responsible for compliant control may be attributed to its simple yet feasible effect similar with an impedance controller, and adding an intrinsic impedance (stiffness) outside the control system with no feedback does not interfere the system's performance, especially in the frequency domain.

### B. Neural Network

With ample data and training, the neural network generates the prediction of ankle outputs, including ankle angle ($\alpha$) and angular velocity ($\dot{\alpha}$). The raw data extracted from the musculoskeletal model and the predicted results are shown in Fig. 5. Each gait is essentially portrayed by the trajectory on the predicted surfaces, allowing continuous recognition and alteration of locomotive speed.

For gait phase prediction of all speeds, 70% of result has a relative error within 5%, while 84% of it has a relative error within 10%. For speed range of 0~2m/s, 43% of the predicted speed has a relative error within 10%, while 76% of it has a relative error within 25%. For speed range of 2~5m/s, 56% of the predicted speed has a relative error within 10%, while 91% of it has a relative error within 25%. The difference between speed intervals may be attributed to the transmission from walking to running, which typically occurs at 2m/s when the motion pattern of lower limb changes significantly [48].

For angle prediction, 41% of the result has a relative error of 10%, while 63% of it has a relative error of 25%. However, 87% of ankle prediction has a relative error within 10% and 96% of it has a relative error of 25%, once the angle is bigger than 15°. The range of angle above 15° accounts for 20%~60% of an entire gait cycle in the speed range of 0~5 m/s. Additionally, deviation of the original data does not influence the prediction result. For example, although the deviation of swing phase is usually smaller than the stance phase in raw data, predictions in both phases appear identical in error yet follow the rule of 15°. This may be attributed to same absolute

TABLE 3 OPTIMIZED PARAMETERS OF THE PLANAR TORSIONAL SPRING

| H | 11.5 mm | $R_{outer}$ | 3.5 mm |
|---|---|---|---|
| h | 10 mm | $T_{inner}$ | 1.2 mm |
| D | 2.4 mm | L | 10 mm |

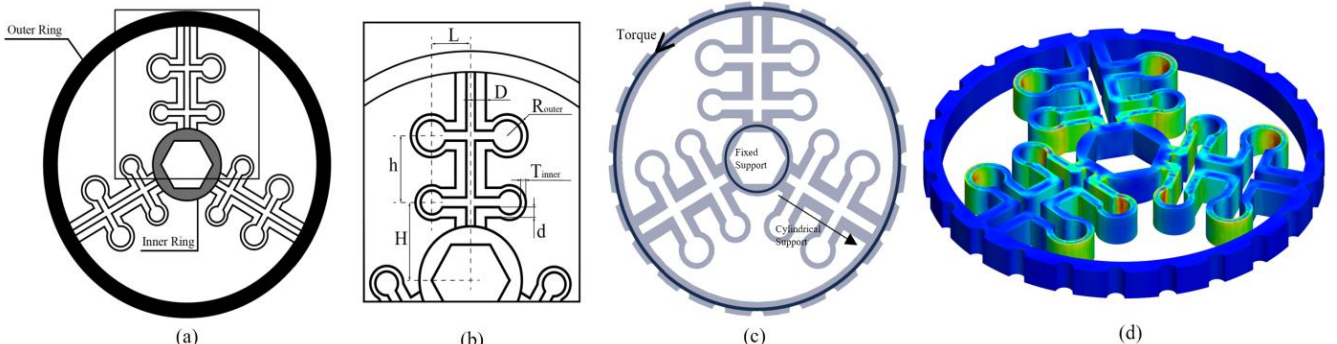

Fig. 4 Planar torsional spring. (a) Design scheme. (b) Partial enlarged view. (c) Load method. (d) Static analysis.

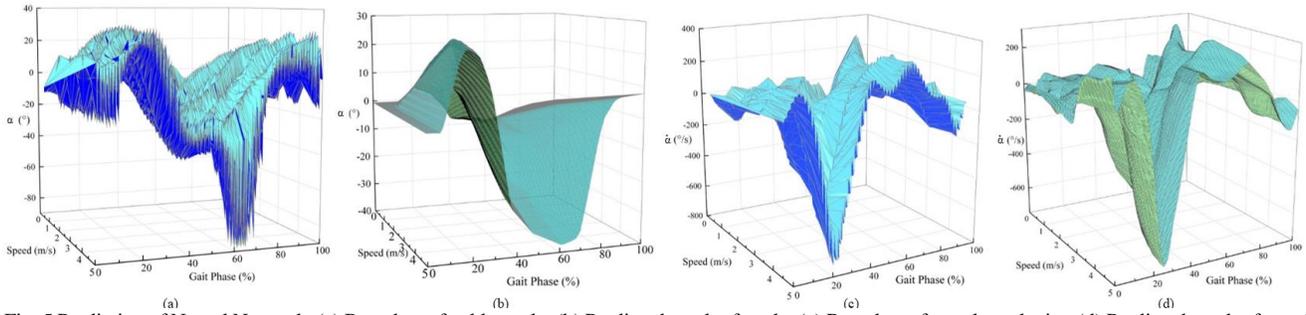

Fig. 5 Prediction of Neural Network. (a) Raw data of ankle angle. (b) Predicted result of angle. (c) Raw data of angular velocity. (d) Predicted result of angular velocity

error turning vastly different when divided by different desired results. Yet when converting the evaluation methodology of the network training to relative error, the situation does not alter much. Error of angle prediction below 15° can be indicated with mean square error of the result, which is approximately 3°. The accuracy is enough for a field prosthesis, since gait data have an intrinsic deviation of a similar level [2]. The prediction of angular velocity is barely satisfactory, with a mean square error of about 50°/s. The inaccuracy of prediction can be attributed to significant deviation of the raw data as well as the ambiguous relationship among angular velocity, speed, and phase. The dramatic variation in amplitude is mostly due to the small angle motion early in stance phase after heel strike, when the joint unsteadily moves in a small range, causing even larger fluctuation for angular velocity. Thus, this variable is only

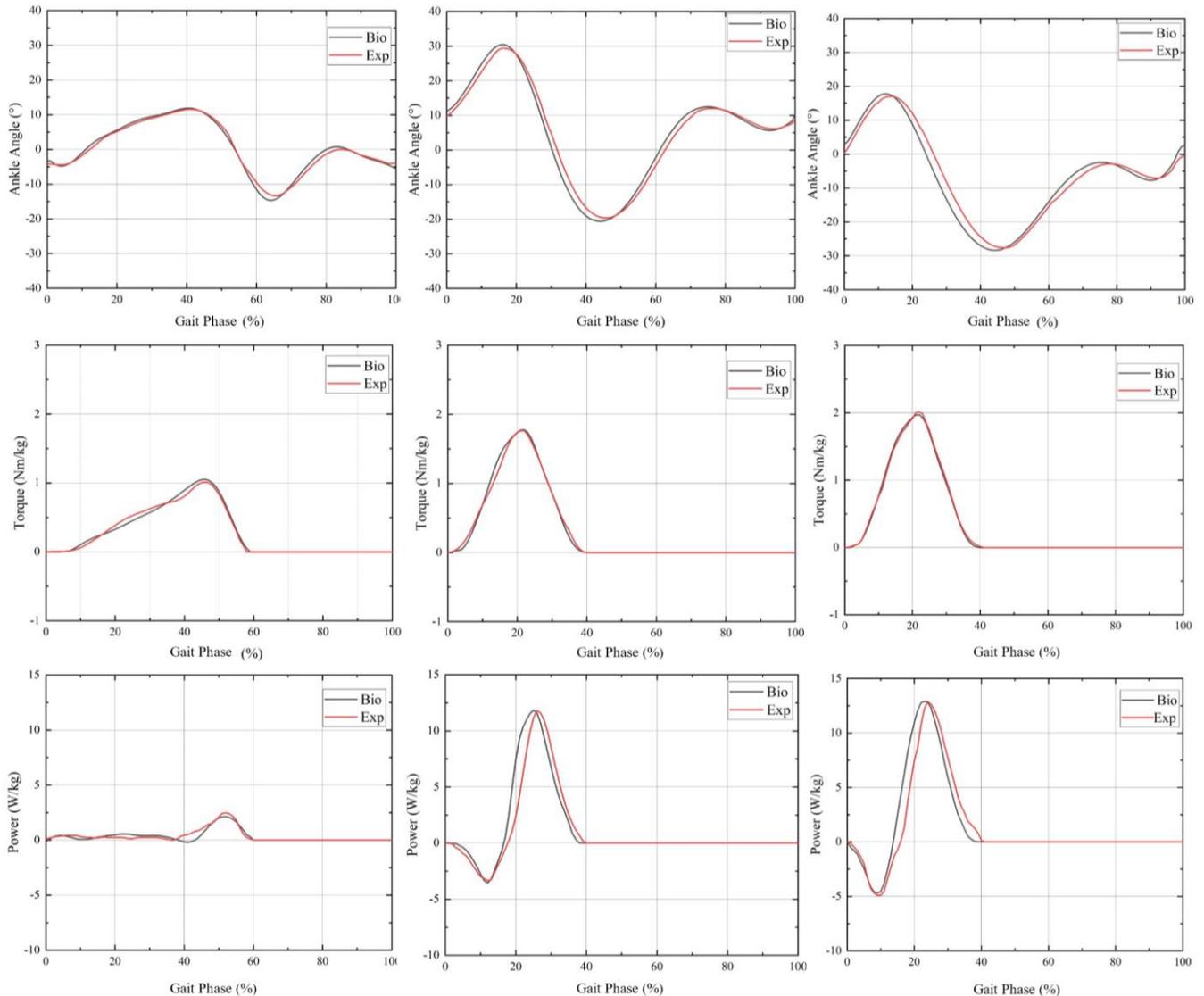

Assuming a transmission of 4.6:1

Fig. 6 Test result of prosthesis. Columns from left to right each illustrates 1.2m/s, 3/2m/s, 3.9m/s. rows from top to bottom each illustrates output angle, torque, and power.

taken into accounts for better understanding of gait dynamics.

For better predictions, a subsequent filter performing hypothesis-testing can be utilized to exclude big error and an average result. Because the inputs and outputs of the system follows Lipschitz condition, when continuous data from IMU generate continuous outputs, the restriction on adjacent output filters out big accidental errors and therefore elevates the accuracy. The frequency of calculation depends mostly on IMU sampling, and is now set at 100Hz. The estimated output frequency above 10Hz is acceptable and achievable.

*C. Prosthesis*

Field tests of the proposed ankle prosthesis were conducted to verify the functionality of the gait recognition and control method. The subject wore an L-shaped socket that attaches to the subject's shank at one end and the prosthesis at the other end (Fig. 1e). The platform enables healthy people without amputation to conduct experiments. The subjects were required to move in three typical speeds for daily life, namely 1.2m/s for walking, 3.2 m/s for running, and 3.9m/s for sprinting [2]. For each group, at least 30 gait cycles are measured. In order to accurately measure the outputs, rigid actuator is adopted instead of SEA. The result is shown in Fig. 6.

Ankle angles in the control group (black line, labeled "Bio") is generated by the neural network, while angles in the experimental group (red line, labeled "Exp") is measured by the motor's encoder. Torque and power in the control group is from previous research [2], while in the experimental group, torque is calculated from the motor's current minus the idle current without load, and power is calculated from torque and rotational speed retrieved by the encoder.

The experimental results show perfect real-time tracking to the biological data. There is a delay of angle for up to 3 % of gait cycle. The torque has an error of up to 3 %, especially during ascending, which may be caused by unstable measurement of the motor's current. In addition, the peak value of angle is 5 % lower than the control group, yet the peak values of output torque and power present to be identical to the control group.

The error of angle's amplitude as well as delay may be caused by the PID controller, the delay of signal transmission, and the systematic error of measurement. The error of torque may be attributed to inaccurate measurement of idle current. In addition, the stance phase produces majority of output torque and power, yet in the swing phase the ankle only rotates without supporting anything. The transition between these two phases is indicated when torque and power returns zero. The major delay, which usually appears when the joint angle in the experimental group descends from its first peak, happens exactly before toe-off, a crucial point after the torque reaches its peak and provides maximum assistance to the human body.

V. CONCLUSION

This paper prosed an innovative powered ankle prosthesis with new design of planar torsional spring for SEA, and utilizes MLP for gait prediction and recognition. The spring has been proved in simulation to have appropriate stiffness and outstanding endurance. With functional relationships between each parameter versus stiffness and maximum von Mises stress verified, theoretical support for modification and optimization is provided. The relation between the shank angles and angular velocities measured by a single IMU are used to establish a phase variable control method that characterizes gait cycle. The network excels in its multi-functional capability in handling continuous prediction of gait phase, locomotive speed, ankle angle, and angular velocity. Though the accuracy is not so competitive compared to previous algorithms specifically for single mission, the range of motion is enhanced to 0~5m/s, bigger than existing gait recognition models and holding outstanding performance compared to other assistive wearable robotics. Field tests also proved its assistive effect of up to 4m/s and mimicking of biological ankle.

Furthermore, this paper verifies the hypothesis that single neural network can accomplish joint works of numerous algorithm each specific for single tasks, which significantly reduces system's complexity. The neural network discovers more of a statistical connection between all selected variables, instead of physical models, such as pendulum for speed prediction. With sufficient data gathered through various means and intricate techniques for optimization, neural networks can handle sophisticated MIMO system in a rather ambiguous yet effective way. Future works of this project will focus on control method for speed change and other daily motions like squatting. Due to the modularity of this joint, a full lower limb prosthesis consisting knee and ankle joints is also planned. With amputees involved in the experiments, the functionality would be testified in a more thorough and clear way.

ACKNOWLEDGMENT

This paper is jointly accomplished by three undergraduate students under the supervision of their tutor. The author would like to extend sincere gratitude to Mr. Wang, Mr. Zhao, Miss Ji, Mr. Wu, and Miss Tang for their assistance.

REFERENCES

[1] K. Ziegler-Graham, E. J. MacKenzie, P. L. Ephraim, T. G. Travison, and R. Brookmeyer, "Estimating the prevalence of limb loss in the United States: 2005 to 2050," Archives of physical medicine and rehabilitation, vol. 89, no.3, pp. 422-429, 2005.

[2] T. F. Novacheck, "The biomechanics of running," Gait & Posture, vol. 7, no. 1, pp. 77-95, 1998.

[3] H. Hsiao, B. A. Knarr, J. S. Higginson, and S. A. Binder-Macleod, "The relative contribution of ankle moment and trailing limb angle to propulsive force durimng gait," Human movement science, vol. 39, pp. 212-221, 2015.

[4] B. J. Hafner, J. E. Sanders, J. M. Czerniecki, and J. Fergason, "Transtibial energy-storage-and-return prosthetic devices: a review of energy concepts and a proposed nomenclature," Journal of Rehabilitation Research & Development, vol. 39, no. 1, 2002.

[5] R. Versluys, P. Beyl, M. Van Damme, A. Desomer, R. Van Ham, and D. Lefeber, "Prosthetic feet: state-of-the-art review and the importance of mimicking human ankle–foot biomechanics," Disability and Rehabilitation: Assistive Technology, vol. 4, no. 2, pp. 65-75, 2009.

[6] P. Cherelle, G. Mathijssen, Q. Wang, B. Vanderborght, and D. Lefeber, "Advances in propulsive bionic feet and their actuation principles," Advances in Mechanical Engineering, vol. 6, p. 984046, 2014.

[7] H. Zheng and X. Shen, "Sleeve muscle actuator and its application in transtibial prostheses," in Proc. of the 2013 IEEE 13th International Conference on Rehabilitation Robotics (ICORR), 2013.

[8] X. Wang, R. Li, J. Fang, S. Wang, and C. Lin, "A powered ankle prothesis driven by EHA technique," in Proc. of the 2018 13th IEEE Conference on Industrial Electronics and Applications (ICIEA), 2018, pp. 1492-1497.

[9] W. Jang, D. Kim, Y. Choi, and Y. Kim, "Self-contained 2-DOF ankle-foot prosthesis with low-inertia extremity for agile walking on uneven terrain," IEEE Robotics and Automation Letters, vol. 6, no. 4, pp. 8134-8141, 2021.

[10] S. Yu, T. Huang, X. Yang, C. Jiao, J. Yang, Y. Chen, J. Yi, and H. Su, "Quasi-direct drive actuation for a lightweight hip exoskeleton with high backdrivability and high bandwidth," IEEE/ASME Transactions on Mechatronics, vol. 25, no. 4, pp. 1794-1802, 2020.

[11] H. Zhao, E. Ambrose, and A. D. Ames, "Preliminary results on energy efficient 3D prosthetic walking with a powered compliant transfemoral prosthesis," in Proc. of the 2017 IEEE International Conference on Robotics and Automation (ICRA), 2017, pp. 1140-1147.

[12] P. Cherelle, V. Grosu, L. Flynn, K. Junius, M. Moltedo, B. Vanderborght, and D. Lefeber, "The Ankle Mimicking Prosthetic Foot 3—Locking mechanisms, actuator design, control and experiments with an amputee," Robotics and Autonomous Systems, Vol. 91, pp. 327-336, 2017.

[13] R. Martinez, R. Avitia, R.; M. Bravo, and M. Reyna, "A low cost design of powered ankle-knee prosthesis for lower limb amputees - preliminary results," in Proc. of the International Conference on Biomedical Electronics and Devices (BIOSTEC 2014), pp. 253-258, 2014.

[14] S. Wolf, G. Grioli, O. Eiberger, W. Friedl, M. Grebenstein, H. Höppner, E. Burdet, D. G. Caldwell, R. Carloni, M. G. Catalano, D. Lefeber, S. Stramigioli, M. Van Damme, R. Van Ham, B. Vanderborght, L. C. Visser, A. Bicchi, and A. Albu-Schäffer, "Variable stiffness actuators: review on design and components," IEEE/ASME Trans. Mechatronics, vol. 21, no. 5, pp. 2418-2430, 2015.

[15] N. Hogan, "Impedance control: an approach to manipulation: part II—implementation," ASME Journal of Dynamic Systems, Measurement, and Control, vol. 107, pp. 8-16, 1985.

[16] Y. Wen, J. Si, A. Brandt, X. Gao and H. H. Huang, "Online Reinforcement Learning Control for the Personalization of a Robotic Knee Prosthesis," IEEE Transactions on Cybernetics, vol. 50, no. 6, pp. 2346-2356, 2020.

[17] S. Song, L. Kidzinski, X. Peng, C. Ong, J. Hicks, S. Levine, C. G. Atkeson, and S. L. Delp, "Deep reinforcement learning for modeling human locomotion control in neuromechanical simulation," Journal of NeuroEngineering and Rehabilitation, vol. 18, no. 126, 2021.

[18] G. Grandesso, G. Bravo-Palacios, P. M. Wensing, M. Fontana, and A. D. Prete, "Exploring the limits of a redundant actuation system through co-design," IEEE Access, vol. 9, pp. 56802-56811, 2021.

[19] I. Kang, H. Hsu and A. Young, "The Effect of Hip Assistance Levels on Human Energetic Cost Using Robotic Hip Exoskeletons," IEEE Robotics and Automation Letters, vol. 4, no. 2, pp. 430-437, 2019.

[20] T. Zhang and H. Huang, "A Lower-Back Robotic Exoskeleton: Industrial Handling Augmentation Used to Provide Spinal Support," IEEE Robotics & Automation Magazine, vol. 25, no. 2, pp. 95-106, 2018.

[21] A. F. Azocar, L. M. Mooney, L. J. Hargrove, and E. J. Rouse, "Design and characterization of an open-source robotic leg prosthesis," in Proc. of the 7th IEEE International Conference on Biomedical Robotics and Biomechatronics (Biorob), pp. 111-118, 2018.

[22] N. G. Tsagarakis, M. Laffranchi, B. Vanderborght, and D. G. Caldwell, "A compact soft actuator unit for small scale human friendly robots," in Proc. of the 2009 International Conference on Robotics and Automation (ICRA), 2009, pp. 4356-4362.

[23] Y. Wang, Y. Chen, K. Chen, Y. Wu, and Y. Huang, "A flat torsional spring with corrugated flexible units for series elastic actuators," in Proc. of the 2017 2nd International Conference on Advanced Robotics and Mechatronics (ICARM), 2017, pp. 138-143.

[24] S. Wang et al., "Design and control of the MINDWALKER exoskeleton," IEEE Transactions on Neural Systems and Rehabilitation Engineering, 2015, vol. 23, no. 2, pp. 277-286.

[25] S. A. Murray, K. H. Ha, C. Hartigan and M. Goldfarb, "An assistive control approach for a lower-limb exoskeleton to facilitate recovery of walking following stroke," IEEE Transactions on Neural Systems and Rehabilitation Engineering, 2015, vol. 23, no. 3, pp. 441-449.

[26] D. J. Villarreal, D. Quintero and R. D. Gregg, "Piecewise and unified phase variables in the control of a powered prosthetic leg," 2017 International Conference on Rehabilitation Robotics (ICORR), 2017, pp. 1425-1430.

[27] K. Seo et al., "RNN-based on-Line continuous gait phase estimation from shank-mounted IMUs to control ankle exoskeletons," 2019 IEEE 16th International Conference on Rehabilitation Robotics (ICORR), 2019, pp. 809-815.

[28] D. Quintero, D. J. Lambert, D. J. Villarreal, and R. D. Gregg, "Real-Time continuous gait phase and speed estimation from a single sensor," 2017 IEEE Conference on Control Technology and Applications (CCTA), 2017, pp. 847-852.

[29] M. Shushtari, R. Nasiri and A. Arami, "Online Reference Trajectory Adaptation: A Personalized Control Strategy for Lower Limb Exoskeletons," IEEE Robotics and Automation Letters, vol. 7, no. 1, pp. 128-134, 2022.

[30] A. F. Azocar, L. M. Mooney, J. F. Duval, A, M. Simon, L. J. Hargrove, and E. J. Rouse, "Design and clinical implementation of an open-source bionic leg," Nature biomedical engineering, vol. 4, no. 10, pp. 941-953, 2020.

[31] M. R. Tucker, J. Olivier, A. Pagel, H. Bleuler, M. Bouri, O. Lambercy, J. R. Millán, R. Riener, H. Vallery, and R. Gassert, "Control strategies for active lower extremity prosthetics and orthotics: a review," Journal of NeuroEngineering and Rehabilitation, vol. 12, no. 1, pp. 1-30, 2015.

[32] U. C. Ugbolue, C. Robson, E. Donald, K. L. Speirs, F. Dutheil, J. S. Baker, T. Dias, and Y. Gu, "Joint angle, range of motion, force, and moment assessment: responses of the lower limb to ankle plantarflexion and dorsiflexion," Bionics and Biomechanics, 2021, pp. 1-13.

[33] P. de Leva, "Adjustments to Zatsiorsky-Seluyanov's segment inertia parameters," Journal of Biomechanics, vol. 29, no. 9, pp. 1223-1230, 1996.

[34] A. Moraux et al., "Ankle dorsi- and plantar-flexion torques measured by dynamometry in healthy subjects from 5 to 80 years," BMC Musculoskeletal Disorders, vol. 14, no. 104, 2013.

[35] B. F. Mentiplay, M. Banky, R. A. Clark, M. B. Kahn, and G. Williams, "Lower limb angular velocity during walking at various speeds," Gait & Posture, vol. 65, pp. 190-196, 2018.

[36] T. Kim, K. Shi, and K. Kong, "A compact transmitted-force-sensing series elastic actuator with optimized planar torsional spring for exoskeletons," in Proc. of the 2020 IEEE/ASME International Conference on Advanced Intelligent Mechatronics (AIM), 2021, pp. 572-577.

[37] L. F. Van Swam, R. M. Pelloux, and N. J. Grant, "Fatigue behavor of maraging steel 300," Metallurgical Transactions A, vol. 6, pp. 45-54, 1975.

[38] C. L. Dembia, N. A. Bianco, A. Falisse, J. L. Hicks, and S. L. Delp, "Opensim moco: musculoskeletal optimal control," PLOS Computational Biology, vol. 16, no.12, p. e1008493, 2020.

[39] A. Seth et al., "Opensim: simulating musculoskeletal dynamics and neuromuscular control to study human and animal movement," PLoS computational biology, vol. 14, no. 7, p. e1006223, 2018.

[40] S. L. Delp et al., "OpenSim: open-source software to create and analyze dynamic simulations of movement," IEEE transactions on biomedical engineering, vol. 54, no. 11, pp. 1940-1950, 2007.

[41] C. John, F. Anderson, J. Higginson, and S. Delp, "Stabilisation of walking by intrinsic muscle properties revealed in a three-dimensional muscle-driven simulation," Computer Methods in Biomechanics and Biomedical Engineering, 2013, vol. 16, no. 4, pp. 451-462.

[42] S. R. Hamner and S. L. Delp, "Muscle contributions to fore-aft and vertical body mass center accelerations over a range of running speeds," Journal of Biomechanics, vol. 46, no. 4, pp. 780-787, 2013.

[43] N. Goldfarb, A. Lewis, A. Tacescu, and G. S. Fischer, "Open source vicon toolkit for motion capture and gait analysis," Computer Methods and Programs in Biomedicine, vol. 212, p. 106414, 2012.

[44] J. M. Franklin, J. A. Rassen, D. Ackermann, D. B. Bartels, and S. Schneeweiss, "Metrics for covariate balance in cohort studies of causal effects," Statistics in Medicine, vol. 33, no. 10, pp. 1685-1699, 2014.

[45] X. Zhao, W. Lin, J. Li, Y. Chen, A. Patel, H. Zhao, G. Han, Y. Hao, C. Fu, Z. Huang, M. Zheng, and P. Hu, "Dose correlation of Danggui and Chuanxiong drug pairs in the Chinese medicine prescription based on the copula function," Evidence-Based Complementary and Alternative Medicine, 2020, pp. 1-10.

[46] M. Elgart et al., "Correlations between complex human phenotypes vary by genetic background, gender, and environment," Cell Reports Medicine, vol. 3, no. 12, p. 100844, 2012.

[47] R. Forthofer, E. Lee, and M. Hernandez, Biostatistics, 2nd ed., chapter 3, Elsevier, 2007, pp. 21-69.

[48] A. Raynor, C. Yi, B. Abernethy, Q. Jong, "Are transitions in human gait determined by mechanical, kinetic or energetic factors," Human Movement Science, 2002, vol. 21, pp. 785-805.